\documentclass{article}

\usepackage[main, final]{neurips_2025}

\usepackage[utf8]{inputenc} 
\usepackage[T1]{fontenc}    
\usepackage{hyperref}       
\usepackage{url}            
\usepackage{booktabs}       
\usepackage{amsfonts}       
\usepackage{nicefrac}       
\usepackage{microtype}      
\usepackage{xcolor}         

\usepackage{amsmath}
\usepackage{amssymb}    
\usepackage{amsthm}     
\usepackage{tcolorbox} 
\usepackage{latexsym}
\usepackage{times}
\usepackage{algorithm}    
\usepackage[noend]{algpseudocode} 
\usepackage{float}        
\usepackage{wrapfig}
\usepackage{lipsum} 

\usepackage{inconsolata}

\usepackage{multirow}
\usepackage{wrapfig}
\usepackage{natbib}
\usepackage{multicol}
\usepackage{tabularx}
\usepackage{graphicx}
\usepackage{pifont}
\usepackage{colortbl}
\usepackage{longtable}
\usepackage{ulem}
\usepackage{listings}
\usepackage{xspace}
\usepackage{fancyvrb}
\usepackage{adjustbox}
\usepackage{tikz}
\usepackage{forest}

\newcommand{\ourmodel}[0]{{\textsc{DynaAct}}\xspace}
\newtheorem{lemma}{Lemma}

\title{\ourmodel: Large Language Model Reasoning with Dynamic Action Spaces}

\author{
\textbf{Xueliang Zhao}$^\spadesuit$$^{\bigstar}$\thanks{\hspace{2mm}This work was done during an internship at Ant Group.} \quad
\textbf{Wei Wu}$^{\bigstar}$\thanks{\hspace{2mm}Corresponding authors.} \quad
\textbf{Jian Guan}$^{\bigstar}$ \quad
\textbf{Qintong Li}$^\spadesuit$ \quad
\textbf{Lingpeng Kong}$^\spadesuit$\footnotemark[2] \\
  $^\spadesuit$The University of Hong Kong \quad
  $^\bigstar$Ant Group \\
\texttt{\{xlzhao,qtli,lpk\}@cs.hku.hk}\\
\texttt{\{wuwei19850318,jianguanthu\}@gmail.com} \\
}

\begin{document}
\maketitle
\begin{abstract}
In modern sequential decision-making systems, the construction of an optimal candidate action space is critical to efficient inference. However, existing approaches either rely on manually defined action spaces that lack scalability or utilize unstructured spaces that render exhaustive search computationally prohibitive. In this paper, we propose a novel framework named \textsc{DynaAct} for automatically constructing a compact action space to enhance sequential reasoning in complex problem-solving scenarios. Our method first estimates a proxy for the complete action space by extracting general sketches observed in a corpus covering diverse complex reasoning problems using large language models. We then formulate a submodular function that jointly evaluates candidate actions based on their utility to the current state and their diversity, and employ a greedy algorithm to select an optimal candidate set. Extensive experiments on six diverse standard benchmarks demonstrate that our approach significantly improves overall performance, while maintaining efficient inference without introducing substantial latency. The implementation is available at \url{https://github.com/zhaoxlpku/DynaAct}.

\end{abstract}
\section{Introduction}
Recent advances in complex reasoning with Large Language Models (LLMs)~\citep{achiam2023gpt,jaech2024openai} have established a prevalent self-improvement paradigm: given a mass of problems paired with final answers, model developers first search for correct reasoning paths from a base model using test time scaling strategies~\citep{snell2024scaling}, then improve it to internalize these patterns~\citep{guo2025deepseek} through supervised fine-tuning~\citep{guan2025rstar} or reinforcement learning~\citep{guo2025deepseek}. While LLMs have exhibited remarkable reasoning capabilities, current approaches to long-term reasoning in these models often suffer from  fundamental limitations.  On the one hand, some approaches explicitly define an action space and state space and structure reasoning hierarchically, where action selection and state prediction are carried out iteratively \citep{hao2023reasoning,qi2024mutual} along the reasoning path. However, because action spaces are often heuristically designed, the resulting actions tend to either be too specific to generalize across domains or too broad to effectively guide reasoning. On the other hand, in the absence of an explicit definition for action spaces, other approaches impose a specific format  on generation and perform reasoning in an autoregressive manner \citep{lightman2023let,guo2025deepseek}. These approaches inherently search the entire natural language space for reasoning, thereby necessitating powerful base models.

We investigate LLM reasoning within the framework of Markov Decision Process (MDP) \citep{hao2023reasoning}, where a reasoning trace consists of a series of actions and states. Instead of focusing on policy learning or reward modeling, we pay special attention to action space construction, as a well-defined action space is fundamental to MDP-based reasoning. Specifically, we identify two essential properties that a qualified action space should possess: \textbf{(1) Scalability:} it should be automatically learned from demonstration data, rather than being manually engineered, to strike a balance between generalization and utility; and \textbf{(2) Compactness:} it should maintain a dynamically constructed, information-dense structure so that for each step desired action can be picked up from a small yet complete candidate set. The core technical challenge lies in developing a principled approach that simultaneously optimizes both objectives: distilling generalizable action patterns from demonstrations while eliminating redundant candidates that would otherwise impede efficient exploration.

To achieve both scalability and compactness, we frame the problem of action space construction as a subset selection task and propose \ourmodel, in which the action space for each reasoning step is dynamically determined by a submodular function that is learned through a data-driven approach. The key idea is to approximate a small subset from the entire action space that achieves the optimal balance between utility and diversity, leveraging the diminishing returns property of submodular functions to ensure linear computational complexity. Our method begins by extracting general reasoning patterns from a diverse corpus of complex problems to construct the complete action space. We then define a submodular function to evaluate candidate actions by jointly considering their utility to the current state and their diversity contribution. By maximizing this function via a greedy algorithm, we obtain an optimal subset of actions. Consequently, reasoning follows a standard Markov process: at each step, candidate actions are selected via the submodular optimization, an action is then chosen according to a Q-function estimated via Monte Carlo tree search, and a reasoning step is finally generated conditioned on the current reasoning context and the chosen action. Notably, throughout this process, only a lightweight embedding model used in the submodular function requires training, while the base LLM remains frozen.

We conduct extensive experiments on six benchmarks spanning general, reasoning, and math tasks. Evaluation results indicate that \ourmodel achieves significant improvements over baselines across all tasks, including MMLU, MMLU-Pro, GPQA, ARC-C, GSM8K, and MATH-500. Notably, \ourmodel excels at solving complex problems, achieving a 6.8\% absolute gain over the recently proposed strong model rStar on MATH-500. Furthermore, an extended study demonstrates that while dynamic action space construction enhances efficacy, it does not introduce significant additional latency during inference compared to the baselines.

Our contributions are three-fold: (1) We propose dynamic action space construction as a novel research question, orthogonal to the extensive studies on LLM reasoning in the community; (2) We introduce a submodular function for action space construction, which significantly improves problem-solving accuracy while maintaining reasonable inference efficiency; and (3) We empirically verify the efficacy of our method across a wide range of tasks.
\begin{figure*}[t]
\centering
\includegraphics[width=1.0\textwidth]{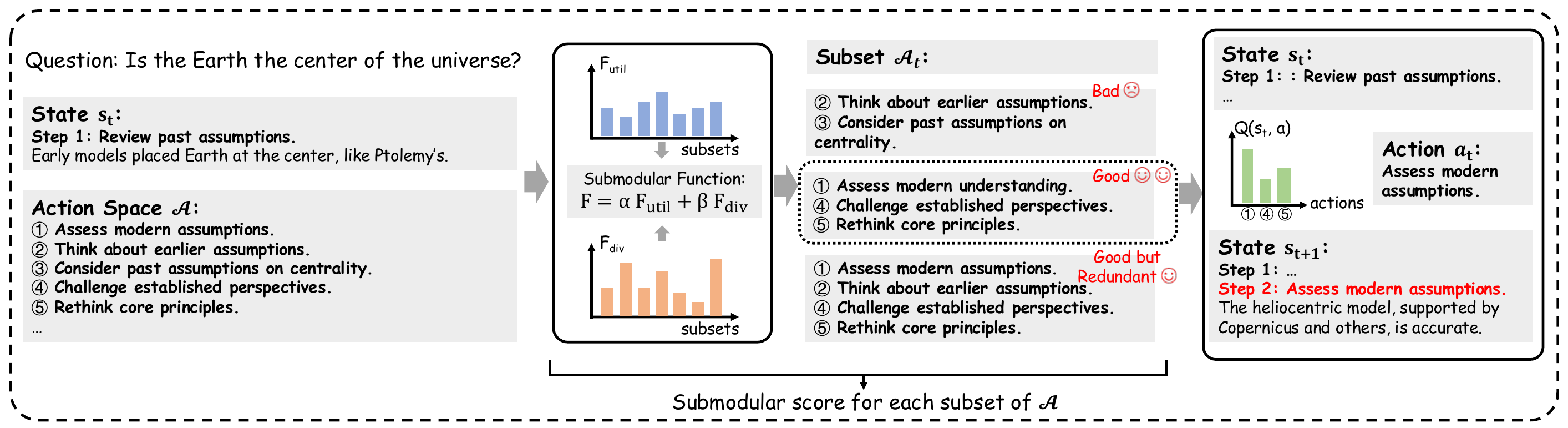}
\vspace{-4mm}
\caption{Overview of the proposed method. Given the proxy action space \(\mathcal{A}\), the method searches for the subset \(\mathcal{A}_t\) that maximizes the submodular function, which consists of a utility term and a diversity term. The subset \(\mathcal{A}_t\) is then used for the subsequent reasoning steps.}
\vspace{-4mm}
\label{fig:method_overview}
\end{figure*}

\section{Preliminaries}
Before delving into \ourmodel, we provide some background information. We begin by formulating MDP-based reasoning, then describe the search strategy applied. The section finally gives a brief introduction to submodular functions, which form the theoretical basis for action space construction.
\subsection{Reasoning Framework}\label{sec:reasoning_framework}
We formulate LLM reasoning as an MDP, defined by a tuple $(\mathcal{S}, \mathcal{A}, \mathcal{T}, \mathcal{R}, \gamma)$. Here, $\mathcal{S}$ represents the state space where each state $s_t \in \mathcal{S}$ encodes the cumulative reasoning context up to step $t$, with the initial state $s_0$ derived from the input prompt. $\mathcal{A}$ denotes the action space with $a_t \in \mathcal{A}$ an action indicating the progression of the reasoning context. The transition function $\mathcal{T}: \mathcal{S} \times \mathcal{A} \rightarrow \mathcal{S}$ determines how actions transform the reasoning state, while $\mathcal{R}: \mathcal{S} \times \mathcal{A} \rightarrow \mathbb{R}$ assigns rewards based on the quality of reasoning steps. The discount factor $\gamma \in [0,1]$ balances immediate and future rewards.

At each time step $t$, the LLM selects an action $a_t$ from a candidate set $\mathcal{A}_t \subseteq \mathcal{A}$, which may be generated automatically~\citep{hao2023reasoning} or specified manually~\citep{qi2024mutual}. 
In our implementation, the action selection is governed by a learnable value function $Q(s_t, a)$ that estimates the expected cumulative reward:
\begin{align*}
    a_t &= \arg\max_{a \in \mathcal{A}_t} Q(s_t, a),\\
    Q(s_t, a) &= \mathbb{E}\left[\sum_{k=0}^{\infty} \gamma^k \mathcal{R}(s_{t+k}, a_{t+k}) \mid s_t, a_t=a\right].
\end{align*}
Our primary focus is on developing a principled approach to construct the candidate action set $\mathcal{A}_t$ at each step, as this significantly impacts the efficiency and efficacy of the reasoning process in complex problem-solving scenarios. 

\subsection{Monte Carlo Tree Search}
We employ Monte Carlo Tree Search (MCTS) for estimating $Q(s_t,a)$~\citep{silver2016mastering}.
In a nutshell, the estimation is achieved by simulating multiple continuations of the reasoning process from the current state after applying the candidate action. During these simulations, the outcomes of the extended reasoning traces are evaluated, and the value of an action is approximated as the average result observed across the simulations. In this way, MCTS effectively balances the exploration of less frequently visited candidate actions with the exploitation of those that have demonstrated promising progress, which is critical to ensure that the estimated value function reliably reflects the potential benefit of each action in guiding the overall reasoning process. 
We defer additional technical details of MCTS, including how selection, expansion, simulation, and backpropagation are performed, to Appendix~\ref{sec:appendix_mcts}.

\subsection{Submodular Functions}

The problem of action space construction can naturally be formalized as a subset selection task, where the goal is to identify a small, high-value subset from a much larger set of potential candidates. A typical approach to subset selection involves assessing the utility of each element, ensuring that every new element added contributes additional value to the overall set. To achieve this, submodular functions are often employed due to their diminishing returns property, which prioritizes the selection of elements that offer unique and informative value \citep{fujishige2005submodular,kothawade2022talisman,chen2024less}. As a result, subset selection with submodular functions ensures that the marginal benefit of adding an element to a smaller subset is greater than adding it to a larger one, thus enhancing the compactness of the subset.

Formally, given two candidate sets \(X \subseteq X'\) and an action \(a \in X \setminus X'\), a submodular function $F(\cdot; \cdot)$ satisfies the following condition:
\begin{equation}\label{eq:submodular_def}
\begin{aligned} 
F(X \cup \{a\}; s_t) - F(X; s_t) \geq F(X' \cup \{a\}; s_t) - F(X'; s_t).
\end{aligned}
\end{equation}
Heading toward a scalable approach for constructing compact action spaces, we take advantage of submodular functions. The problem then boils down to (1) how to define a proper submodular function; (2) how to learn the submodular function from data; and (3) how to perform action space construction with the submodular function, as will be presented in the following Section. 

\begin{algorithm*}
\small
\caption{Complete Pipeline of Sequential Reasoning.}\label{alg:full_pipeline}
\begin{algorithmic}[1]
\Require Input question \(q\), dataset \(\mathcal{D}\), number of groups \(k\), candidate selection budget \(m\), maximum reasoning steps \(T\)
\State {/* Proxy Action Space Estimation (performed once) */}
\State Partition the dataset \(\mathcal{D}\) into \(k\) groups: \(\{\mathcal{D}_1, \mathcal{D}_2, \dots, \mathcal{D}_k\}\)
\For{\(i = 1\) to \(k\)}
    \State Extract an observation sketch \(o_i = \left\langle a_1, a_2, \dots, a_{|o_i|} \right\rangle \gets \text{LLMQuery}(\mathcal{D}_i)\)
\EndFor
\State Form the action space \(\mathcal{A} = \bigcup_{i=1}^{k} o_i = \bigcup_{i=1}^{k} \left\langle a_1, a_2, \dots, a_{|o_i|} \right\rangle\)

\State Train the embedding function \(\mathbf{e}\) using Q-learning objective (Eq.~\eqref{eq:sqil}) with observation sketches as demonstration data

\State \(s_0 \gets \text{InitializeState}(q)\)
\For{\(t = 0\) to \(T - 1\)}

    \State {/* Candidate Action Selection via Greedy Algorithm */}
    \State \(X \gets \emptyset\)
    \For{\(i = 1\) to \(m\)}
        \State \(X_d \gets \mathcal{A} \setminus X\)
        \State \(a^* \gets \arg\max_{a \in X_d} F(X \cup \{a\}; s_t)\)
        \State \(X \gets X \cup \{a^*\}\)
    \EndFor
    \State \(\mathcal{A}_t(s_t) \gets X\)
    
    \State {/* Action Evaluation using MCTS */}
    \ForAll{\(a \in \mathcal{A}_t(s_t)\)}
        \State \(Q(s_{t}, a) \gets \text{MCTS}(s_{t}, a)\)
    \EndFor
    
    \State {/* Action Selection and State Update */}
    \State \(a_t \gets \arg\max_{a \in \mathcal{A}_t(s_t)} Q(s_{t-1}, a)\)
    \State \(s_{t+1} \gets \text{UpdateState}(s_{t}, a_t)\)
\EndFor
\Statex \hspace*{-\algorithmicindent} \textbf{Output:} \(\{s_0, a_0, s_1, \dots, s_T\}\)
\end{algorithmic}
\end{algorithm*}

\section{Method}
We detail the approach to constructing $\mathcal{A}_t$ given the current state \(s_t\). Specifically, our method consists of three stages: estimating an approximation of the complete action space as $\mathcal{A}$~(\S\ref{initialize}), defining a submodular function $F(\mathcal{A}_t, s_t)$ based on \(s_t\)~(\S\ref{submodular}), and utilizing the function to derive $\mathcal{A}_t$~(\S\ref{selection}). Figure~\ref{fig:method_overview} provides an overview of the method.

\subsection{Proxy Action Space Estimation}\label{initialize}
We first estimate an approximation as a proxy of the complete action space, denoted as \(\mathcal{A}\). Specifically, we follow \citet{wang2024planning} by employing observations as candidate actions, where observations are typically cues that guide the reasoning process (cf. Figure~\ref{fig:method_overview}).
Given a problem corpus (e.g., mathematical questions, logical puzzles, etc.), we randomly divide it into \(k\) groups and feed each group to an LLM for observation collection. The division strategy ensures that each group is appropriately sized, avoiding prohibitive computational costs from the LLM. We then query the LLM to extract general observation sketchs per group that can be applied broadly and focus solely on the core operations. After that, the resulting observations are collected and form the proxy action space \(\mathcal{A}\) with duplicate items removed. The prompt for observation sketch extraction is provided in Appendix~\ref{sec:appendix_sketch}. Notably, $\mathcal{A}$ can be easily scaled up by incorporating examples from broader domains or fields; one can also develop various agents by applying the method to domain-specific corpora. 

\subsection{Submodular Function Definition}\label{submodular}

To construct an optimal candidate action set that balances both utility (in terms of expected rewards) and diversity, we propose a submodular function $F(\mathcal{A}_t; s_t)$ for a candidate subset $\mathcal{A}_t \subseteq \mathcal{A}$ as follows:
\begin{equation}\label{eq:submodular}
F(\mathcal{A}_t; s_t) = \alpha\, f_{\mathrm{util}}(\mathcal{A}_t; s_t) + \beta\, f_{\mathrm{div}}(\mathcal{A}_t),
\end{equation}
where $f_{\mathrm{util}}(\mathcal{A}_t; s_t)$ measures the expected utility of the candidate actions in advancing the reasoning process, $f_{\mathrm{div}}(\mathcal{A}_t)$ promotes diversity within the selected set, and $\alpha, \beta$ are balancing parameters. To ensure that $F(\mathcal{A}_t; s_t)$ defined by Eq.~\eqref{eq:submodular} meets the condition given by Eq.~\eqref{eq:submodular_def}, we define the utility term as:
\begin{equation}\label{eq:submodular_util}
f_{\mathrm{util}}(\mathcal{A}_t; s_t) = \log \left( \sum_{a \in \mathcal{A}_t} \exp\Bigl( \mathbf{e}(s_t)^T \mathbf{e}(a)\Bigr) \right),
\end{equation}
where $\mathbf{e}(\cdot)$ is an embedding function that maps states and actions to a shared representation space. Then the diversity term is defined as
\begin{equation}\label{eq:submodular_div}
f_{\mathrm{div}}(\mathcal{A}_t) = \sum_{a_i \in \mathcal{A}_t} \min_{\substack{a_j \in \mathcal{A}_t \\ a_j \neq a_i}} \Bigl(1 - \mathbf{e}(a_i)^T \mathbf{e}(a_j)\Bigr),
\end{equation}
This formulation encourages the selection of actions that are maximally distinct from each other in the embedding space, preventing redundancy in the candidate set. 
\begin{lemma}\label{lem}
Given the definitions of the relevance term \(f_{\mathrm{util}}(\mathcal{A}_t; s_t)\) in Eq.~\eqref{eq:submodular_util} and the diversity term \(f_{\mathrm{div}}(\mathcal{A}_t)\) in Eq.~\eqref{eq:submodular_div}, the function \(F(\mathcal{A}_t; s_t)\) defined in Eq.~\eqref{eq:submodular} is submodular with respect to the candidate action set \(\mathcal{A}_t\subseteq \mathcal{A}\).
\end{lemma}
The proof of Lemma~\ref{lem} is provided in Appendix~\ref{sec:appendix_submodular_proof}.

A fundamental requirement of our framework is ensuring that $\mathcal{A}_t$ contains actions that maximize expected rewards in the reasoning process. To this end, we design the embedding function $\mathbf{e}(\cdot)$ to capture the effectiveness of actions in advancing the reasoning process. We formalize this requirement through Q-learning, where $\mathbf{e}(s_t)^T \mathbf{e}(a)$ approximates the Q-value—the expected future reward of executing action $a$ in state $s_t$. By incorporating this formulation into the standard Q-learning update equation~\citep{watkins1992q, reddy2019sqil}, we derive the following optimization objective:
\begin{equation}
\begin{aligned}\label{eq:sqil}
&\mathcal{L}(s_t, a, s_{t+1}) = \mathbf{e}(s_t)^T\, \mathbf{e}(a) - \left( r + \log \left( \sum_{a' \in \mathcal{A}} \exp\left( \mathbf{e}(s_{t+1})^T\, \mathbf{e}(a') \right) \right) \right)^2,
\end{aligned}
\end{equation}
where the training data consists of state-action pairs $(s_t, a_t)$, with $a_t \in \mathcal{A}$ being the ground-truth action obtained from the observation sketches. The reward $r$ is defined as $r = 1$ for $a = a_t$ and $r = 0$ for all other actions $a \in \mathcal{A}\setminus\{a_t\}$. We train the embedding model by minimizing $\mathcal{L}$ over all states $s_t$ in the training set and all actions $a \in \mathcal{A}$. This objective ensures that the embedding function learns to prioritize actions that contribute substantively to the problem-solving progression.

\subsection{Action Space Construction}\label{selection}
With $\mathcal{A}$ and  \(F(\cdot; \cdot)\) at hand, we aim to derive $\mathcal{A}_t$ by selecting \(m\) elements from $\mathcal{A}$ that maximize \(F(\cdot; s_t)\). Formally, we construct $\mathcal{A}_t$ by solving

\begin{equation}\label{eq:combprob}
\mathcal{A}_t(s_t) = \arg\max_{X \subseteq \mathcal{A},\, |X| = m} F(X; s_t).
\end{equation}
Owing to the submodularity of \(F(\cdot; \cdot)\), the combinatorial optimization problem (Eq.~\eqref{eq:combprob}) can be effectively approximated by a greedy algorithm~\citep{nemhauser1978analysis} with complexity of \(O(m^2 |\mathcal{A}|)\), where \(|\mathcal{A}|\) denotes the size of $\mathcal{A}$. In practice, we begin with an empty set \(X\) and iteratively add the element from \(\mathcal{A}\setminus X\) that produces the highest marginal increase in \(F(\cdot; \cdot)\) when combined with the current set \(X\). Formally, for each iteration, we define the set of remaining candidates \(X_d = \mathcal{A}\setminus X\) and select the element \(a\) such that
\[
a = \arg\max_{a \in X_d} F(X \cup \{a\}; s_t).
\]
The process is repeated until \(m\) elements have been selected, and the final set \(X\) is then taken as \(\mathcal{A}_t\). 

Algorithm~\ref{alg:full_pipeline} summarizes the complete pipeline of our reasoning method, encompassing both action space construction and MCTS-based reasoning path search.

\vspace{-1mm}
\section{Experiments}
\subsection{Benchmarks}
We employ six standard benchmarks covering three domains: general, reasoning, and math. Specifically, we use the following datasets:

(1) \textbf{MMLU}~\citep{hendrycks2020measuring} is a benchmark designed to evaluate a model's ability to answer a wide variety of tasks, including reading comprehension, reasoning, and problem-solving, across general domains. It is widely used for assessing language model performance in broad tasks;
(2) \textbf{MMLU-Pro}~\citep{wang2024mmlu} is an extension of MMLU, containing more challenging and professional-level problems. This dataset tests the model's capabilities on more complex problems in general domains;
(3) \textbf{GPQA}~\citep{rein2023gpqa} focuses on evaluating reasoning and problem-solving skills, providing real-world open-domain problems that require advanced reasoning to solve;
(4) \textbf{ARC-challenge (ARC-C)}~\citep{clark2018think} is part of the AI2 Reasoning Challenge and contains science-based multiple-choice questions. These questions require deep reasoning and are specifically designed to challenge models in complex reasoning tasks;
(5) \textbf{GSM8K}~\citep{cobbe2021training} is a dataset consisting of grade-school level math word problems that require logical reasoning. It tests a model’s ability to solve elementary-level math problems;
and (6) \textbf{MATH-500}~\citep{lightman2023let} is a dataset containing high school-level math problems. It serves to assess a model's ability to handle more advanced mathematical reasoning.

\vspace{-1mm}
\subsection{Evaluation Metrics}

We use exact match accuracy as the primary metric for evaluating the performance of our method. Specifically, for multiple-choice question-answering tasks, such as MMLU, MMLU-Pro, GPQA, and ARC-C, accuracy is calculated based on the exact match between the predicted choice and the ground-truth choice (typically denoted by a letter representing the correct answer).
For math problem-solving tasks, such as GSM8K and MATH-500, accuracy is calculated by comparing the predicted final answer, enclosed by \texttt{\textbackslash boxed\{\}}, with the ground-truth answer.

\vspace{-1mm}
\subsection{Baseline Methods}

We compare the proposed method with the following baselines: (1) \textbf{Zero-shot CoT}: This baseline uses Llama 3.1~\citep{dubey2024llama} with zero-shot Chain-of-Thought (CoT) prompting~\citep{wei2022chain}, generating reasoning paths in a single pass; (2) \textbf{SC@maj16}: This method applies the self-consistency (SC) technique~\citep{wang2022self}, where multiple reasoning paths are generated and the most frequent result is selected. We use the SC@maj16 variant, which runs $16$ rollouts to increase the accuracy of reasoning; (3) \textbf{RAP}: The method~\citep{hao2023reasoning} integrates a world model and a reasoning agent, balancing exploration and exploitation to efficiently find high-reward reasoning paths via MCTS. The action space is formed by automatically generated sub-questions; and (4) \textbf{rStar}: The method~\citep{qi2024mutual} utilizes $5$ manually defined actions as the action space. Reasoning traces are searched with MCTS rollouts, and the final trace is determined by a small LLM as a discriminator.

All of these baselines use Llama-3.1-8B-Instruct~\citep{dubey2024llama} as the backbone, consistent with our method. Additionally, both RAP and rStar use $16$ rollouts, as in our method. 

\vspace{-1mm}
\subsection{Implementation Details}

For the proxy action space estimation~(\S\ref{initialize}), we use the Open-Platypus~\citep{lee2023platypus} corpus, which covers a wide range of topics, including math, scientific reasoning, and more. The corpus contains a total of \(24,652\) problems, which are used to form the problem set. We divide the corpus into \(k = 2,500\) groups. To extract observations, we query Llama-3.1-70B-Instruct~\citep{dubey2024llama}, resulting in \(40,822\) observations in total. 
In our submodular function definition~(\S\ref{submodular}), we set the balancing parameters \(\alpha = 0.9\) and \(\beta = 0.1\) to ensure a proper balance between utility and diversity. For embedding efficiency, we use Llama-3.2-1B-Instruct~\citep{dubey2024llama} as the backbone and select the last token's embedding as the output of the embedding function \(\mathbf{e}(\cdot)\). The embedding function is fine-tuned using the Q-learning objective, with a total of \(83,083\) state-action pairs, and the learning rate is set to \(1e-5\).
For the action space construction~(\S\ref{selection}), we set the size of the candidate action set \(\mathcal{A}_t\) at each time step \(m = 5\). 
We use Llama-3.1-8B-Instruct~\citep{dubey2024llama} as the world model~\citep{hao2023reasoning}, which generates reasoning steps during the MCTS process.

\begin{table*}[!ht]
\centering
\vspace{-2mm}
\caption{Evaluation results on different benchmarks. Numbers in bold denote the best performance.}
\label{tab:performance}
\begin{tabular}{l@{\hspace{0.5em}}cc@{\hspace{1em}}cc@{\hspace{1em}}cc}
\toprule
& \multicolumn{2}{c}{General} & \multicolumn{2}{c}{Reasoning} & \multicolumn{2}{c}{Math} \\
\cmidrule(lr){2-3} \cmidrule(lr){4-5} \cmidrule(lr){6-7}
Model & MMLU & MMLU-Pro & GPQA & ARC-C & GSM8K & MATH-500 \\
\midrule
Zero-shot CoT & 68.87 & 43.45 & 31.82 & 81.06 & 76.12 & 45.40 \\
SC@maj16 & 69.66 & 49.36 & 34.34 & 80.63 & 86.66 & 52.00 \\
RAP & 69.46 & 48.70 & 38.89 & 85.41 & 87.79 & 51.60 \\
rStar & 68.61 & 48.81 & 36.87 & 86.43 & 87.11 & 54.20 \\
\midrule
\ourmodel & \textbf{70.22} & \textbf{51.40} & \textbf{39.39} & \textbf{88.31} & \textbf{89.16} & \textbf{61.00} \\
\bottomrule
\end{tabular}
\end{table*}

\subsection{Main Results}

The results of our experiments, presented in Table~\ref{tab:performance}, reveal several key observations:
(1) Our method outperforms all baseline methods across the six evaluated benchmarks, achieving significant improvements in general, reasoning, and math tasks. This confirms the effectiveness of our method in a wide range of problem-solving tasks;
(2) In the math domain, we observe the most notable improvements. Our method achieves 1.37\% and 6.80\% improvements over the baselines on GSM8K and MATH-500, respectively. The MATH-500 dataset, which requires a higher level of reasoning capability, particularly benefits from the use of the submodular function. The ability to construct a more compact and utility-optimized action space enables more efficient exploration, leading to improved performance in complex mathematical reasoning tasks;
and (3) While rStar performs better than RAP on MATH-500, its performance in other benchmarks suffers due to the limited scalability of its manually defined action space. This limits rStar's ability to effectively handle the full range of tasks, making it less scalable compared to our method.

\section{Discussions}

In addition to the comprehensive evaluation across multiple benchmarks, we aim to dive deeper into \ourmodel to gain further insights into its underlying mechanisms. Specifically, we investigate the following research questions: (1) \textbf{RQ1:} How do different components affect performance? (2) \textbf{RQ2:} Can \ourmodel learn to have a compact action space, thereby facilitating efficiency in reasoning? (3) \textbf{RQ3:} What is the utility of the actions selected by \ourmodel? (4) \textbf{RQ4:} Does \ourmodel introduce additional latency during inference?  Besides, we are also curious about (5) \textbf{RQ5:} How does reasoning performance vary across different levels of difficulty? and (6) \textbf{RQ6:} Can the submodular function enhance diversity of actions? We leave the discussions to \textbf{RQ5} and \textbf{RQ6} to Appendix~\ref{sec:appendix_exp}.

\subsection{Ablation Study for RQ1}\label{sec:ablation}
\begin{wraptable}{r}{0.5\textwidth}
\vspace{-10pt}  
\centering
\caption{Ablation study.}
\label{tab:ablation}
\begin{tabular}{l@{\hspace{0.5em}}cc}
\toprule
Model & ARC-C & MATH-500 \\
\midrule
\ourmodel (full) & \textbf{88.31} & \textbf{61.00} \\ \midrule
- util & 87.63 & 53.40 \\
- div & 86.52 & 53.80 \\
- q-learning & 87.80 & 55.80 \\ 
- submodular & 85.15 & 52.00 \\
\bottomrule
\end{tabular}
\vspace{-10pt}
\end{wraptable}

We exploit  ARC-C and MATH-500 to strike a balance between difficulty and domain diversity, and examine four variants of \ourmodel, including:
exclusion of the utility term, denoted as ``- util''; exclusion of the diversity term, referred to as ``- div''; submodular function without Q-learning defined by Eq.~\eqref{eq:sqil}, denoted as ``- q-learning'', which uses Llama-3.2-1B-Instruct for embedding directly; and removal of the submodular function, represented as ``- submodular'', which generates action spaces using Llama-3.1-8B-Instruct directly. 

The results shown in Table \ref{tab:ablation} indicate that the full version of \ourmodel achieves the best performance across both benchmarks, underscoring the critical importance of each component in our method. Removing the utility term leads to a slight decrease in performance, highlighting its contribution to overall reasoning effectiveness. Excluding the diversity term results in further performance degradation, emphasizing the need for diversity in the candidate action set. Omission of the Q-learning objective causes a noticeable drop in performance, demonstrating the necessity of this learning procedure for constructing an effective submodular function. Finally, the model without the submodular function performs the worst, reinforcing the essential role of the submodular strategy in achieving effective action selection.

\subsection{Compactness Study for RQ2}\label{sec:compact}

\begin{figure}[!ht]
    \centering
    \includegraphics[width=0.6\columnwidth]{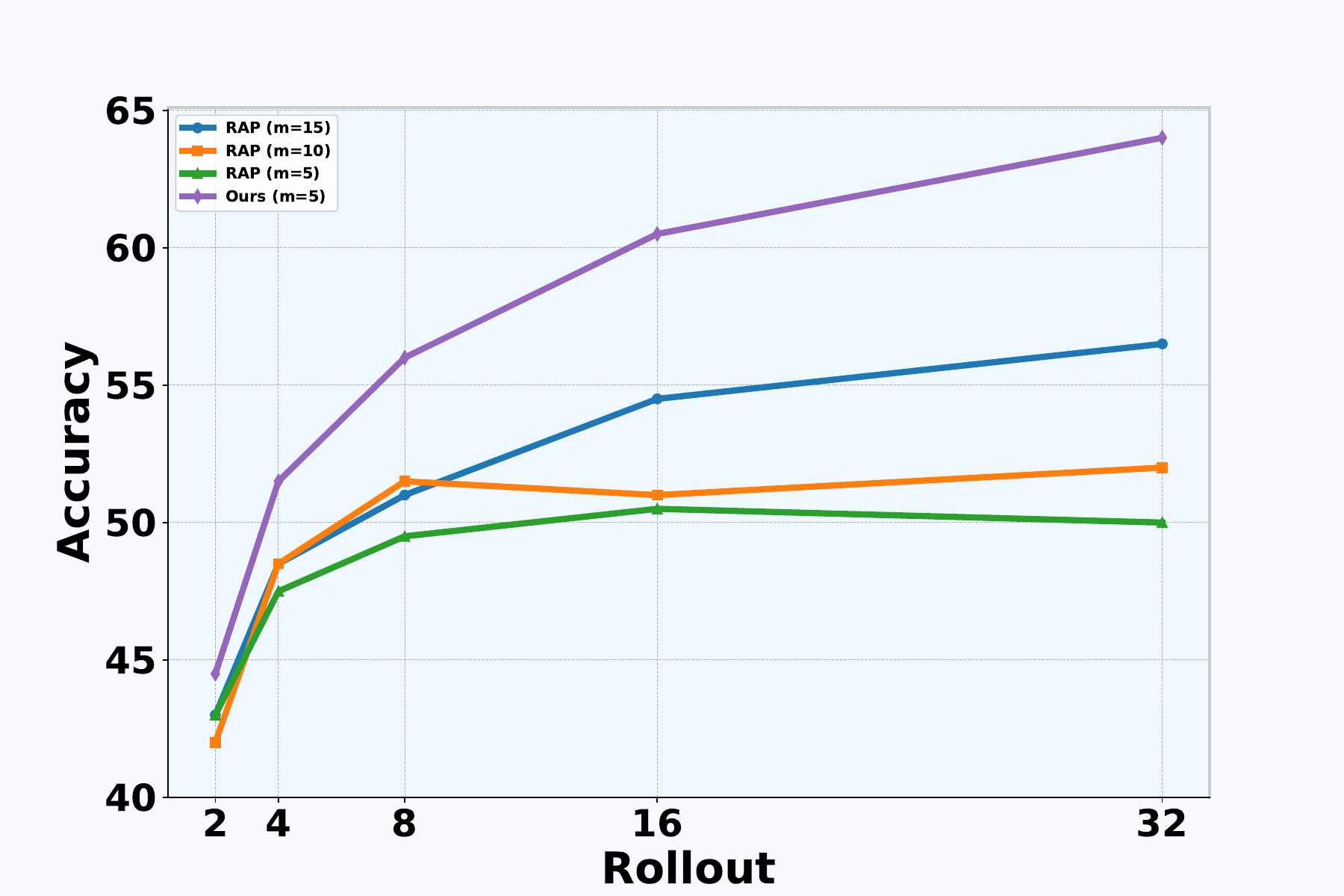} 
    \caption{A comparison of RAP and \ourmodel with respect to different action space sizes (i.e., $m$). The $x$-axis indicates the number of rollouts, while the $y$-axis shows the accuracy on MATH-500.}
    \vspace{-2mm}
    \label{fig:compactness_results} 
    \vspace{-2mm}
\end{figure}

We examine whether \ourmodel can produce compact action spaces. 
Ideally, a compact action space would enhance search efficiency, leading to better reasoning performance with a smaller size. Figure~\ref{fig:compactness_results} compares \ourmodel with RAP where the actions are constrained to be sub-questions. From the results, we observe that: (1) The action spaces of RAP are highly redundant. Increasing the number of rollouts yields limited performance improvements when $m=5$ or $m=10$. Only when $m$ is increased to $15$ do we observe noticeable performance gains with respect to the number of rollouts, though the marginal gains are still much slower compared to \ourmodel. (2) On the other hand, even with $m$ set to $5$, \ourmodel still demonstrates significant performance improvements as more rollouts are carried out. The advantages over RAP remain consistent across all values of $m$, highlighting the efficacy of \ourmodel in generating compact action spaces.

\subsection{Utility Study for RQ3}

\begin{wraptable}{r}{0.60\textwidth}
\vspace{-10pt}
\centering
\caption{Evaluation results on the utility of action spaces. ``Accuracy'' is measured on a subset of MATH-500 with Level 5 difficulty (i.e., the most difficult subset). This subset is chosen because the ``utility'' of actions (i.e., if critical steps are triggered) plays a more critical role in solving complex problems.}
\label{tab:scalability_results}
\begin{tabular}{lcc}
\toprule
Model & Critical Step Coverage (\(\uparrow\)) & Accuracy (\(\uparrow\)) \\
\midrule
\ourmodel & 0.63 & 31.34 \\
rStar     & 0.47 & 26.87 \\
\bottomrule
\end{tabular}
\vspace{-10pt}
\end{wraptable}

We select rStar as the baseline since its manually designed actions are presumed to be overly broad, limiting their utility for reasoning. Typically, there is no standard method for measuring ``utility''. Therefore, following \citet{zhao2024bba}, we identify critical steps in problem-solving (e.g., key insights, decisions, or calculations essential for solving a problem) and examine whether the reasoning steps triggered by actions contain these critical steps. The ratio of solutions containing critical steps (referred to as the \textit{Critical Step Coverage}) is then used as a proxy metric for utility. Table~\ref{tab:scalability_results} presents the evaluation results. We can see that \ourmodel enables more effective identification and triggering of essential reasoning steps, which also explains its superiority over rStar in terms of accuracy.

\subsection{Latency Study for RQ4}\label{sec_latency_study}

\begin{wraptable}{r}{0.55\textwidth}
\vspace{-10pt}
\centering
\caption{Comparison of relative time and accuracy for different methods.}
\label{tab:latency}
\begin{tabular}{lcc}
\toprule
Method & Rel. Time (\(\downarrow\)) & Accuracy (\(\uparrow\)) \\
\midrule
\ourmodel & 1.00 & 61.00 \\
rStar     & 0.95 & 54.20 \\
RAP       & 1.12 & 51.60 \\
\bottomrule
\end{tabular}
\vspace{-10pt}
\end{wraptable}

To assess whether \ourmodel introduces additional latency during inference, we compare it with rStar and RAP based on the relative time required to complete tasks on the MATH-500 dataset. Specifically, relative time is calculated as the time taken by rStar or RAP, relative to the time taken by \ourmodel. We set the action space size (denoted as $m$) to $5$ and the number of rollouts to $16$ for all methods. Table~\ref{tab:latency} presents the results, including accuracy on MATH-500 to examine the latency-accuracy trade-off. The results show that while \ourmodel incurs a slight increase in latency compared to rStar, it achieves substantial improvements in accuracy. When compared to RAP, our method demonstrates lower latency, which can be attributed to the fact that the most computationally intensive operation in constructing the action space $\mathcal{A}_t$ is the encoding of $\mathbf{e}(s_t)$, with $\mathbf{e}(a)$ precomputed and cached for subsequent use. This contrasts with RAP, where sub-questions are generated in real-time, leading to higher computational overhead. Additionally, the submodular function can be computed efficiently using an approximate algorithm with linear complexity, further contributing to the reduced latency of our approach.

\vspace{-1mm}
\section{Related Work}
\vspace{-1mm}
\subsection{LLM Reasoning} 
The exploration of LLMs' reasoning capabilities has emerged alongside studies of prompting strategies, among which \citet{wei2022chain} demonstrated that a simple prompt can elicit chain-of-thought reasoning in LLMs; \citet{zhouleast} leveraged prompts to guide LLMs in breaking down complex problems into simpler ones; and \citet{shinn2024reflexion} enabled LLMs to recognize their mistakes and perform self-correction.
Soon after, the research on reasoning shifted focus from prompting to data curation \citep{zhao2025promptcot,zhao2025promptcot2} and learning methods \citep{zelikman2022star,zelikman2024quiet}, whereby the community witnessed rapid progress in tackling complex problems such as mathematics \citep{yu2023metamath,gou2023tora,mitra2024orca,toshniwal2024openmathinstruct}, coding \citep{luowizardcoder}, visual comprehension \citep{hu2024visual}, and decision-making \citep{chen2023fireact}. Recently, the success of OpenAI o1 \citep{jaech2024openai} and DeepSeek r1 \citep{guo2025deepseek} has catalyzed the rise of test-time scaling \citep{snell2024scaling}, where increased computation at inference enables LLMs to engage in long-term reasoning and achieve significant improvements on Olympiad-level math \citep{aimedata}, challenging coding benchmarks \citep{jain2024livecodebench}, and graduate-level QA tasks \citep{rein2023gpqa}. 
Our study contributes to test-time scaling research but takes an orthogonal approach to most existing efforts. Rather than focusing on data curation \citep{muennighoff2025s1,guan2025rstar} or reinforcement learning \citep{guo2025deepseek,qi2024mutual}, we aim to construct a compact yet effective action space to facilitate MDP-based reasoning. 

\vspace{-1mm}
\subsection{Submodular Optimization}
Submodular optimization~\citep{fujishige2005submodular} has been successfully used in many applications, including model interpretability~\citep{elenberg2017streaming,chen2018learning,chen2024less} and computer vision~\citep{pervez2023scalable}. For instance, 
\citet{elenberg2017streaming} applied submodular functions to model interpretability, framing it as a combinatorial maximization problem for efficient model explanations. Similarly, \citet{chen2018learning} used submodular functions for instance-wise feature selection to explain deep learning decisions. \citet{pervez2023scalable} introduced conditional Poisson sampling to select key features for image and text recognition, focusing on improving recognition accuracy.
More recently, \citet{chen2024less} applied submodular subset selection to deep model attribution, enhancing interpretability by identifying critical regions and reducing misattribution.
Our work applies submodular optimization to action space selection in sequential reasoning tasks. Unlike prior methods focused on feature selection and model attribution, our approach optimizes action selection for enhanced compactness and scalability, offering a novel method that improves performance in complex problem-solving.
\section{Conclusion}
We propose \ourmodel, a novel approach for automatically constructing compact action spaces to enhance sequential reasoning in complex problem-solving tasks. \ourmodel incorporates a submodular function that optimizes action selection based on both utility and diversity, leading to improved inference efficiency and performance. Extensive experiments across six standard benchmarks demonstrate that \ourmodel not only outperforms existing methods but also maintains efficient inference without introducing substantial latency. These results underscore the effectiveness and versatility of our approach in tackling a wide range of problem solving challenges.

\section*{Acknowledgements}
We extend our gratitude to the HKU NLP group and the anonymous reviewers for their invaluable suggestions, which significantly enhanced this work.
This work was supported in part by the joint research scheme of the National Natural Science Foundation of China (NSFC) and the Research Grants Council (RGC) under grant number N\_HKU714/21, and by the Ant Group Research Intern Program.

\bibliographystyle{plainnat}
\bibliography{custom}

\clearpage
\section*{NeurIPS Paper Checklist}

\begin{enumerate}

\item {\bf Claims}
    \item[] Question: Do the main claims made in the abstract and introduction accurately reflect the paper's contributions and scope?
    \item[] Answer: \answerYes{} 
    \item[] Justification: The abstract and introduction accurately reflect the paper's contributions and scope.
    \item[] Guidelines:
    \begin{itemize}
        \item The answer NA means that the abstract and introduction do not include the claims made in the paper.
        \item The abstract and/or introduction should clearly state the claims made, including the contributions made in the paper and important assumptions and limitations. A No or NA answer to this question will not be perceived well by the reviewers. 
        \item The claims made should match theoretical and experimental results, and reflect how much the results can be expected to generalize to other settings. 
        \item It is fine to include aspirational goals as motivation as long as it is clear that these goals are not attained by the paper. 
    \end{itemize}

\item {\bf Limitations}
    \item[] Question: Does the paper discuss the limitations of the work performed by the authors?
    \item[] Answer: \answerYes{} 
    \item[] Justification: Details are discussed in the Limitations section.
    \item[] Guidelines:
    \begin{itemize}
        \item The answer NA means that the paper has no limitation while the answer No means that the paper has limitations, but those are not discussed in the paper. 
        \item The authors are encouraged to create a separate "Limitations" section in their paper.
        \item The paper should point out any strong assumptions and how robust the results are to violations of these assumptions (e.g., independence assumptions, noiseless settings, model well-specification, asymptotic approximations only holding locally). The authors should reflect on how these assumptions might be violated in practice and what the implications would be.
        \item The authors should reflect on the scope of the claims made, e.g., if the approach was only tested on a few datasets or with a few runs. In general, empirical results often depend on implicit assumptions, which should be articulated.
        \item The authors should reflect on the factors that influence the performance of the approach. For example, a facial recognition algorithm may perform poorly when image resolution is low or images are taken in low lighting. Or a speech-to-text system might not be used reliably to provide closed captions for online lectures because it fails to handle technical jargon.
        \item The authors should discuss the computational efficiency of the proposed algorithms and how they scale with dataset size.
        \item If applicable, the authors should discuss possible limitations of their approach to address problems of privacy and fairness.
        \item While the authors might fear that complete honesty about limitations might be used by reviewers as grounds for rejection, a worse outcome might be that reviewers discover limitations that aren't acknowledged in the paper. The authors should use their best judgment and recognize that individual actions in favor of transparency play an important role in developing norms that preserve the integrity of the community. Reviewers will be specifically instructed to not penalize honesty concerning limitations.
    \end{itemize}

\item {\bf Theory assumptions and proofs}
    \item[] Question: For each theoretical result, does the paper provide the full set of assumptions and a complete (and correct) proof?
    \item[] Answer: \answerYes{} 
    \item[] Justification: Details are included in Appendix E.
    \item[] Guidelines:
    \begin{itemize}
        \item The answer NA means that the paper does not include theoretical results. 
        \item All the theorems, formulas, and proofs in the paper should be numbered and cross-referenced.
        \item All assumptions should be clearly stated or referenced in the statement of any theorems.
        \item The proofs can either appear in the main paper or the supplemental material, but if they appear in the supplemental material, the authors are encouraged to provide a short proof sketch to provide intuition. 
        \item Inversely, any informal proof provided in the core of the paper should be complemented by formal proofs provided in appendix or supplemental material.
        \item Theorems and Lemmas that the proof relies upon should be properly referenced. 
    \end{itemize}

    \item {\bf Experimental result reproducibility}
    \item[] Question: Does the paper fully disclose all the information needed to reproduce the main experimental results of the paper to the extent that it affects the main claims and/or conclusions of the paper (regardless of whether the code and data are provided or not)?
    \item[] Answer: \answerYes{} 
    \item[] Justification: Details are discussed in Section 4.4.
    \item[] Guidelines:
    \begin{itemize}
        \item The answer NA means that the paper does not include experiments.
        \item If the paper includes experiments, a No answer to this question will not be perceived well by the reviewers: Making the paper reproducible is important, regardless of whether the code and data are provided or not.
        \item If the contribution is a dataset and/or model, the authors should describe the steps taken to make their results reproducible or verifiable. 
        \item Depending on the contribution, reproducibility can be accomplished in various ways. For example, if the contribution is a novel architecture, describing the architecture fully might suffice, or if the contribution is a specific model and empirical evaluation, it may be necessary to either make it possible for others to replicate the model with the same dataset, or provide access to the model. In general. releasing code and data is often one good way to accomplish this, but reproducibility can also be provided via detailed instructions for how to replicate the results, access to a hosted model (e.g., in the case of a large language model), releasing of a model checkpoint, or other means that are appropriate to the research performed.
        \item While NeurIPS does not require releasing code, the conference does require all submissions to provide some reasonable avenue for reproducibility, which may depend on the nature of the contribution. For example
        \begin{enumerate}
            \item If the contribution is primarily a new algorithm, the paper should make it clear how to reproduce that algorithm.
            \item If the contribution is primarily a new model architecture, the paper should describe the architecture clearly and fully.
            \item If the contribution is a new model (e.g., a large language model), then there should either be a way to access this model for reproducing the results or a way to reproduce the model (e.g., with an open-source dataset or instructions for how to construct the dataset).
            \item We recognize that reproducibility may be tricky in some cases, in which case authors are welcome to describe the particular way they provide for reproducibility. In the case of closed-source models, it may be that access to the model is limited in some way (e.g., to registered users), but it should be possible for other researchers to have some path to reproducing or verifying the results.
        \end{enumerate}
    \end{itemize}

\item {\bf Open access to data and code}
    \item[] Question: Does the paper provide open access to the data and code, with sufficient instructions to faithfully reproduce the main experimental results, as described in supplemental material?
    \item[] Answer: \answerYes{} 
    \item[] Justification: We have submitted the source code to facilitate the reproduction of our results.
    \item[] Guidelines:
    \begin{itemize}
        \item The answer NA means that paper does not include experiments requiring code.
        \item Please see the NeurIPS code and data submission guidelines (\url{https://nips.cc/public/guides/CodeSubmissionPolicy}) for more details.
        \item While we encourage the release of code and data, we understand that this might not be possible, so “No” is an acceptable answer. Papers cannot be rejected simply for not including code, unless this is central to the contribution (e.g., for a new open-source benchmark).
        \item The instructions should contain the exact command and environment needed to run to reproduce the results. See the NeurIPS code and data submission guidelines (\url{https://nips.cc/public/guides/CodeSubmissionPolicy}) for more details.
        \item The authors should provide instructions on data access and preparation, including how to access the raw data, preprocessed data, intermediate data, and generated data, etc.
        \item The authors should provide scripts to reproduce all experimental results for the new proposed method and baselines. If only a subset of experiments are reproducible, they should state which ones are omitted from the script and why.
        \item At submission time, to preserve anonymity, the authors should release anonymized versions (if applicable).
        \item Providing as much information as possible in supplemental material (appended to the paper) is recommended, but including URLs to data and code is permitted.
    \end{itemize}

\item {\bf Experimental setting/details}
    \item[] Question: Does the paper specify all the training and test details (e.g., data splits, hyperparameters, how they were chosen, type of optimizer, etc.) necessary to understand the results?
    \item[] Answer: \answerYes{} 
    \item[] Justification: Details are discussed in Section 4.4.
    \item[] Guidelines:
    \begin{itemize}
        \item The answer NA means that the paper does not include experiments.
        \item The experimental setting should be presented in the core of the paper to a level of detail that is necessary to appreciate the results and make sense of them.
        \item The full details can be provided either with the code, in appendix, or as supplemental material.
    \end{itemize}

\item {\bf Experiment statistical significance}
    \item[] Question: Does the paper report error bars suitably and correctly defined or other appropriate information about the statistical significance of the experiments?
    \item[] Answer: \answerNo{} 
    \item[] Justification: We do not report error bars due to the substantial computational cost associated with repeated runs.
    \item[] Guidelines:
    \begin{itemize}
        \item The answer NA means that the paper does not include experiments.
        \item The authors should answer "Yes" if the results are accompanied by error bars, confidence intervals, or statistical significance tests, at least for the experiments that support the main claims of the paper.
        \item The factors of variability that the error bars are capturing should be clearly stated (for example, train/test split, initialization, random drawing of some parameter, or overall run with given experimental conditions).
        \item The method for calculating the error bars should be explained (closed form formula, call to a library function, bootstrap, etc.)
        \item The assumptions made should be given (e.g., Normally distributed errors).
        \item It should be clear whether the error bar is the standard deviation or the standard error of the mean.
        \item It is OK to report 1-sigma error bars, but one should state it. The authors should preferably report a 2-sigma error bar than state that they have a 96\% CI, if the hypothesis of Normality of errors is not verified.
        \item For asymmetric distributions, the authors should be careful not to show in tables or figures symmetric error bars that would yield results that are out of range (e.g. negative error rates).
        \item If error bars are reported in tables or plots, The authors should explain in the text how they were calculated and reference the corresponding figures or tables in the text.
    \end{itemize}

\item {\bf Experiments compute resources}
    \item[] Question: For each experiment, does the paper provide sufficient information on the computer resources (type of compute workers, memory, time of execution) needed to reproduce the experiments?
    \item[] Answer: \answerYes{} 
    \item[] Justification: Details are discussed in Section 5.4.
    \item[] Guidelines:
    \begin{itemize}
        \item The answer NA means that the paper does not include experiments.
        \item The paper should indicate the type of compute workers CPU or GPU, internal cluster, or cloud provider, including relevant memory and storage.
        \item The paper should provide the amount of compute required for each of the individual experimental runs as well as estimate the total compute. 
        \item The paper should disclose whether the full research project required more compute than the experiments reported in the paper (e.g., preliminary or failed experiments that didn't make it into the paper). 
    \end{itemize}
    
\item {\bf Code of ethics}
    \item[] Question: Does the research conducted in the paper conform, in every respect, with the NeurIPS Code of Ethics \url{https://neurips.cc/public/EthicsGuidelines}?
    \item[] Answer: \answerYes{} 
    \item[] Justification: Yes, the research adheres fully to the NeurIPS Code of Ethics.
    \item[] Guidelines:
    \begin{itemize}
        \item The answer NA means that the authors have not reviewed the NeurIPS Code of Ethics.
        \item If the authors answer No, they should explain the special circumstances that require a deviation from the Code of Ethics.
        \item The authors should make sure to preserve anonymity (e.g., if there is a special consideration due to laws or regulations in their jurisdiction).
    \end{itemize}

\item {\bf Broader impacts}
    \item[] Question: Does the paper discuss both potential positive societal impacts and negative societal impacts of the work performed?
    \item[] Answer: \answerYes{} 
    \item[] Justification: Details are discussed in the Broader Impacts section.
    \item[] Guidelines:
    \begin{itemize}
        \item The answer NA means that there is no societal impact of the work performed.
        \item If the authors answer NA or No, they should explain why their work has no societal impact or why the paper does not address societal impact.
        \item Examples of negative societal impacts include potential malicious or unintended uses (e.g., disinformation, generating fake profiles, surveillance), fairness considerations (e.g., deployment of technologies that could make decisions that unfairly impact specific groups), privacy considerations, and security considerations.
        \item The conference expects that many papers will be foundational research and not tied to particular applications, let alone deployments. However, if there is a direct path to any negative applications, the authors should point it out. For example, it is legitimate to point out that an improvement in the quality of generative models could be used to generate deepfakes for disinformation. On the other hand, it is not needed to point out that a generic algorithm for optimizing neural networks could enable people to train models that generate Deepfakes faster.
        \item The authors should consider possible harms that could arise when the technology is being used as intended and functioning correctly, harms that could arise when the technology is being used as intended but gives incorrect results, and harms following from (intentional or unintentional) misuse of the technology.
        \item If there are negative societal impacts, the authors could also discuss possible mitigation strategies (e.g., gated release of models, providing defenses in addition to attacks, mechanisms for monitoring misuse, mechanisms to monitor how a system learns from feedback over time, improving the efficiency and accessibility of ML).
    \end{itemize}
    
\item {\bf Safeguards}
    \item[] Question: Does the paper describe safeguards that have been put in place for responsible release of data or models that have a high risk for misuse (e.g., pretrained language models, image generators, or scraped datasets)?
    \item[] Answer: \answerNA{} 
    \item[] Justification: The paper does not involve the release of data or models with high risk for misuse.
    \item[] Guidelines:
    \begin{itemize}
        \item The answer NA means that the paper poses no such risks.
        \item Released models that have a high risk for misuse or dual-use should be released with necessary safeguards to allow for controlled use of the model, for example by requiring that users adhere to usage guidelines or restrictions to access the model or implementing safety filters. 
        \item Datasets that have been scraped from the Internet could pose safety risks. The authors should describe how they avoided releasing unsafe images.
        \item We recognize that providing effective safeguards is challenging, and many papers do not require this, but we encourage authors to take this into account and make a best faith effort.
    \end{itemize}

\item {\bf Licenses for existing assets}
    \item[] Question: Are the creators or original owners of assets (e.g., code, data, models), used in the paper, properly credited and are the license and terms of use explicitly mentioned and properly respected?
    \item[] Answer: \answerYes{} 
    \item[] Justification: Details are discussed in Section 4.4.
    \item[] Guidelines:
    \begin{itemize}
        \item The answer NA means that the paper does not use existing assets.
        \item The authors should cite the original paper that produced the code package or dataset.
        \item The authors should state which version of the asset is used and, if possible, include a URL.
        \item The name of the license (e.g., CC-BY 4.0) should be included for each asset.
        \item For scraped data from a particular source (e.g., website), the copyright and terms of service of that source should be provided.
        \item If assets are released, the license, copyright information, and terms of use in the package should be provided. For popular datasets, \url{paperswithcode.com/datasets} has curated licenses for some datasets. Their licensing guide can help determine the license of a dataset.
        \item For existing datasets that are re-packaged, both the original license and the license of the derived asset (if it has changed) should be provided.
        \item If this information is not available online, the authors are encouraged to reach out to the asset's creators.
    \end{itemize}

\item {\bf New assets}
    \item[] Question: Are new assets introduced in the paper well documented and is the documentation provided alongside the assets?
    \item[] Answer: \answerNA{} 
    \item[] Justification: The paper does not release new assets.
    \item[] Guidelines:
    \begin{itemize}
        \item The answer NA means that the paper does not release new assets.
        \item Researchers should communicate the details of the dataset/code/model as part of their submissions via structured templates. This includes details about training, license, limitations, etc. 
        \item The paper should discuss whether and how consent was obtained from people whose asset is used.
        \item At submission time, remember to anonymize your assets (if applicable). You can either create an anonymized URL or include an anonymized zip file.
    \end{itemize}

\item {\bf Crowdsourcing and research with human subjects}
    \item[] Question: For crowdsourcing experiments and research with human subjects, does the paper include the full text of instructions given to participants and screenshots, if applicable, as well as details about compensation (if any)? 
    \item[] Answer: \answerNA{} 
    \item[] Justification: The paper does not involve crowdsourcing nor research with human subjects.
    \item[] Guidelines:
    \begin{itemize}
        \item The answer NA means that the paper does not involve crowdsourcing nor research with human subjects.
        \item Including this information in the supplemental material is fine, but if the main contribution of the paper involves human subjects, then as much detail as possible should be included in the main paper. 
        \item According to the NeurIPS Code of Ethics, workers involved in data collection, curation, or other labor should be paid at least the minimum wage in the country of the data collector. 
    \end{itemize}

\item {\bf Institutional review board (IRB) approvals or equivalent for research with human subjects}
    \item[] Question: Does the paper describe potential risks incurred by study participants, whether such risks were disclosed to the subjects, and whether Institutional Review Board (IRB) approvals (or an equivalent approval/review based on the requirements of your country or institution) were obtained?
    \item[] Answer: \answerNA{} 
    \item[] Justification: The paper does not involve crowdsourcing nor research with human subjects.
    \item[] Guidelines:
    \begin{itemize}
        \item The answer NA means that the paper does not involve crowdsourcing nor research with human subjects.
        \item Depending on the country in which research is conducted, IRB approval (or equivalent) may be required for any human subjects research. If you obtained IRB approval, you should clearly state this in the paper. 
        \item We recognize that the procedures for this may vary significantly between institutions and locations, and we expect authors to adhere to the NeurIPS Code of Ethics and the guidelines for their institution. 
        \item For initial submissions, do not include any information that would break anonymity (if applicable), such as the institution conducting the review.
    \end{itemize}

\item {\bf Declaration of LLM usage}
    \item[] Question: Does the paper describe the usage of LLMs if it is an important, original, or non-standard component of the core methods in this research? Note that if the LLM is used only for writing, editing, or formatting purposes and does not impact the core methodology, scientific rigorousness, or originality of the research, declaration is not required.
    \item[] Answer: \answerNA{} 
    \item[] Justification: The core method does not rely on LLMs for any essential components.
    \item[] Guidelines:
    \begin{itemize}
        \item The answer NA means that the core method development in this research does not involve LLMs as any important, original, or non-standard components.
        \item Please refer to our LLM policy (\url{https://neurips.cc/Conferences/2025/LLM}) for what should or should not be described.
    \end{itemize}

\end{enumerate}
\clearpage
\appendix

\section{Limitations}

While \ourmodel represents a significant advancement in action space construction for reasoning tasks, there are several limitations that should be addressed in future work to fully realize its potential:

(1) \ourmodel demonstrates strong performance across general, reasoning, and math tasks. However, its reliance on MCTS for selecting specific actions from the constructed action space can be computationally intensive, especially for large or complex tasks. Alternative test-time scaling methods, such as beam search or other exploration strategies, could be explored to further optimize the balance between exploration and exploitation, improving efficiency without sacrificing reasoning accuracy. Expanding the model’s ability to scale across different search algorithms remains a promising direction for future work.

(2) While \ourmodel improves reasoning, it still depends on pre-trained Llama models, which may face challenges when handling extremely large or highly specialized action spaces. Exploring methods to optimize the model’s scalability, such as combining \ourmodel with a stronger backbone model, could help tackle this limitation in future research.

\section{Broader Impacts}

This work proposes a framework for dynamic action space construction to improve sequential reasoning with language models. By enabling more structured, efficient, and interpretable decision-making, our approach has the potential to benefit applications such as educational tutoring systems, automated theorem proving, and scientific problem-solving—domains where reasoning efficiency and transparency are critical. The ability to distill compact, high-utility action spaces may also reduce reliance on brute-force search and large-scale inference, contributing to more sustainable and accessible AI systems.

At the same time, these benefits raise important concerns. Automatically constructed action spaces are learned from data and may reflect biases present in the underlying corpora, potentially reinforcing harmful patterns during reasoning. Furthermore, as our method enhances the ability of LLMs to perform multi-step reasoning, it may be misused to generate misleading arguments or automate complex forms of manipulation. While the explicit structure introduced by our framework offers interpretability benefits, future work should explore mechanisms for auditing, constraint enforcement, and safe deployment in sensitive domains.

\section{Details of the MCTS Process}
\label{sec:appendix_mcts}
In this appendix, we provide a detailed formal description of the Monte Carlo Tree Search (MCTS) procedure used to estimate the value function \( Q(s, a) \) for candidate actions. The MCTS process comprises four stages: selection, expansion, simulation, and backpropagation. For simplicity, let the current state be denoted by \( s \), and let the set of candidate actions from state \( s \) be represented as \( \mathcal{A}(s) \).

\subsection*{Selection}
Starting from the root node corresponding to the current state \( s \), the selection phase traverses the tree by recursively choosing actions based on a balance between exploitation and exploration. Each node in the search tree is associated with two statistics: 
\[
N(s) \quad \text{(the number of visits to state } s\text{)}
\]
and, for each action \( a \in \mathcal{A}(s) \),
\[
\begin{array}{rl}
N(s, a) & \text{is the number of times that action $a$ has been selected from state $s$,} \\
Q(s, a) & \text{is the estimated average reward for taking action $a$ from state $s$.}
\end{array}
\]
The action \( a^* \) is selected at state \( s \) according to the Upper Confidence Bound for Trees (UCT) formula:
\begin{align*}
a^* = \arg\max_{a \in \mathcal{A}(s)} \left[ Q(s, a) + c \sqrt{\frac{\ln N(s)}{N(s, a)}} \right],
\end{align*}
where \( c \) is an exploration constant. This selection continues recursively until a node is reached that is either a terminal state or not fully expanded (i.e., there exists an \( a \in \mathcal{A}(s) \) such that the corresponding child node has not yet been created).

\subsection*{Expansion}
Upon reaching a node \( s \) that is not terminal and has unexplored actions, the expansion phase selects one such action \( a' \in \mathcal{A}(s) \) for which no child node exists. A new child node \( s' \) is then created to represent the state resulting from taking action \( a' \) from \( s \). The statistics for the new edge \((s, a')\) are initialized as:
\begin{align*}
N(s, a') &= 0, \\
Q(s, a') &= 0.
\end{align*}

\subsection*{Simulation}
From the newly expanded node \( s' \), a simulation (or rollout) is conducted to estimate the value of the state-action pair \( (s, a') \). The simulation proceeds by following a default policy—often a random or heuristic strategy—to generate a complete reasoning trajectory until a terminal state is reached. Let \( r \) denote the reward obtained at the terminal state; this reward serves as an estimate for the value of taking action \( a' \) from state \( s \).

\subsection*{Backpropagation}
Once the simulation concludes with reward \( r \), the backpropagation phase updates the statistics along the path from the expanded node back to the root. For every state \( s \) and action \( a \) along this path, the updates are performed as follows:
\begin{align*}
N(s, a) &\leftarrow N(s, a) + 1, \\
Q(s, a) &\leftarrow Q(s, a) + \frac{r - Q(s, a)}{N(s, a)}.
\end{align*}
Additionally, the visit count for each state along the path is updated:
\begin{align*}
N(s) \leftarrow N(s) + 1.
\end{align*}
These updates refine the estimates \( Q(s, a) \) based on the observed reward \( r \), thereby improving the accuracy of the value function with successive simulations.

The MCTS process iterates through these steps—selection, expansion, simulation, and backpropagation—until a specified computational budget is reached.

\section{Prompt for Universal Problem-Solving Sketch Extraction}
\label{sec:appendix_sketch}

Below is the prompt used for extracting a universal problem-solving sketch from a set of problems. This prompt is designed to guide the extraction process so that the generated subgoals are broadly applicable, capture core actions, and can be flexibly applied across various problem domains.

\bigskip

\begin{tcolorbox}[colback=blue!5!white, colframe=blue!75!black, title=Problem-Solving Sketch Prompt]
Imagine you are a problem-solving expert tasked with creating a universal problem-solving sketch.

You will be shown the following \{number of problems\} problems:

\{problem\_text\}

From these problems, extract up to \{n\} essential subgoals that form a universal sketch. The subgoals should:
\begin{itemize}
    \item Apply broadly across different types of problems and disciplines
    \item Use casual, everyday language from the first person perspective
    \item Avoid sequential markers like "first", "next", "then", "finally"
    \item Focus on the core action or insight needed at each stage
    \item Be as creative as possible, going beyond what you think is intuitively correct
\end{itemize}

Format your response as a numbered list, where each item expresses one subgoal. Make each subgoal self-contained so it can be applied flexibly rather than in a fixed sequence.
\end{tcolorbox}

\bigskip

This prompt ensures that the extracted subgoals capture the essential observations needed for constructing the proxy action space, and that they are effective for a wide range of problem-solving scenarios.
\section{Proof of Submodularity of \(F(X; s_t)\)}
\label{sec:appendix_submodular_proof}
\begin{proof}
Let \(V = \mathcal{A}\). A set function \(f: 2^V \to \mathbb{R}\) is submodular if for every \(X \subseteq Y \subseteq V\) and for every \(x \in V \setminus Y\),
\[
f(X\cup\{x\}) - f(X) \ge f(Y\cup\{x\}) - f(Y).
\]
We write
\[
F(X; s_t) = \alpha\, F_1(X) + \beta\, F_2(X),
\]
where
\[
F_1(X)=\log\left( \sum_{a\in X}\exp\big(\mathbf{e}(s_t)^T\,\mathbf{e}(a)\big) \right),
\]
and
\[
F_2(X)=\sum_{a\in X} m_X(a), \quad \text{with} \quad m_X(a)=\min_{b\in X\setminus\{a\}} \left( 1 - \mathbf{e}(a)^T \mathbf{e}(b) \right).
\]

\textbf{(i) Submodularity of \(F_1\):}  
Define 
\[
g(X)=\sum_{a\in X}\exp\big(\mathbf{e}(s_t)^T\,\mathbf{e}(a)\big).
\]
Note that for any \(x\in V\setminus X\),
\[
g(X\cup\{x\}) = g(X) + \exp\big(\mathbf{e}(s_t)^T\,\mathbf{e}(x)\big).
\]
Thus, for \(X\subseteq Y\) and \(x\in V\setminus Y\), we have:
\[
F_1(X\cup\{x\}) - F_1(X) = \log\left(1 + \frac{\exp\big(\mathbf{e}(s_t)^T\,\mathbf{e}(x)\big)}{g(X)}\right),
\]
and
\[
F_1(Y\cup\{x\}) - F_1(Y) = \log\left(1 + \frac{\exp\big(\mathbf{e}(s_t)^T\,\mathbf{e}(x)\big)}{g(Y)}\right).
\]
Since \(X\subseteq Y\) implies \(g(X) \le g(Y)\), it follows that
\[
\frac{\exp\big(\mathbf{e}(s_t)^T\,\mathbf{e}(x)\big)}{g(X)} \ge \frac{\exp\big(\mathbf{e}(s_t)^T\,\mathbf{e}(x)\big)}{g(Y)},
\]
and because \(\log(1+z)\) is an increasing function for \(z\ge0\), we obtain
\[
F_1(X\cup\{x\}) - F_1(X) \ge F_1(Y\cup\{x\}) - F_1(Y).
\]
Thus, \(F_1\) is submodular.

\textbf{(ii) Submodularity of \(F_2\):}  
For any \(X\subseteq V\) and \(x\in V\setminus X\), the marginal gain for \(F_2\) is given by
\[
\Delta_{F_2}(x\mid X) \triangleq F_2(X\cup\{x\}) - F_2(X).
\]
For each \(a\in X\), the value \(m_X(a)=\min_{b\in X\setminus\{a\}}\left( 1 - \mathbf{e}(a)^T \mathbf{e}(b) \right)\) updates upon addition of \(x\) to
\[
m_{X\cup\{x\}}(a)=\min\Big\{ m_X(a),\, \left( 1 - \mathbf{e}(a)^T \mathbf{e}(x) \right) \Big\}.
\]
Similarly, for the newly added element \(x\),
\[
m_{X\cup\{x\}}(x)= \min_{a\in X} \left( 1 - \mathbf{e}(x)^T \mathbf{e}(a) \right).
\]
Hence,
\[
\Delta_{F_2}(x\mid X)= \sum_{a\in X}\Big[ \min\{ m_X(a),\, \left( 1 - \mathbf{e}(a)^T \mathbf{e}(x) \right) \} - m_X(a)\Big] + \min_{a\in X}\left( 1 - \mathbf{e}(x)^T \mathbf{e}(a) \right).
\]
Let us denote, for each \(a\in X\),
\[
\delta_X(a,x)= \min\{ m_X(a),\, \left( 1 - \mathbf{e}(a)^T \mathbf{e}(x) \right) \} - m_X(a).
\]
Then,
\[
\Delta_{F_2}(x\mid X)= \sum_{a\in X}\delta_X(a,x) + \min_{a\in X}\left( 1 - \mathbf{e}(x)^T \mathbf{e}(a) \right).
\]
Now, consider \(X\subseteq Y \subseteq V\) and \(x\in V\setminus Y\). For any \(a\in X\), since \(X\subseteq Y\) we have
\[
m_Y(a)= \min\Big\{ m_X(a),\, \min_{b\in Y\setminus X} \left( 1 - \mathbf{e}(a)^T \mathbf{e}(b) \right) \Big\} \le m_X(a).
\]
Thus,
\[
\delta_Y(a,x)= \min\{ m_Y(a),\, \left( 1 - \mathbf{e}(a)^T \mathbf{e}(x) \right) \} - m_Y(a) \ge \min\{ m_X(a),\, \left( 1 - \mathbf{e}(a)^T \mathbf{e}(x) \right) \} - m_X(a) = \delta_X(a,x).
\]
Also,
\[
\min_{a\in Y}\left( 1 - \mathbf{e}(x)^T \mathbf{e}(a) \right) \le \min_{a\in X}\left( 1 - \mathbf{e}(x)^T \mathbf{e}(a) \right).
\]
Therefore,
\begin{align*}
\Delta_{F_2}(x\mid X) &= \sum_{a\in X}\delta_X(a,x) + \min_{a\in X}\left( 1 - \mathbf{e}(x)^T \mathbf{e}(a) \right),\\[1ex]
\Delta_{F_2}(x\mid Y) &= \sum_{a\in Y}\delta_Y(a,x) + \min_{a\in Y}\left( 1 - \mathbf{e}(x)^T \mathbf{e}(a) \right).
\end{align*}
Since \(Y\) contains all elements of \(X\) and possibly additional elements with non-positive marginal increments (i.e., \(\delta_Y(a,x)\le 0\) for \(a\in Y\setminus X\)), it follows that
\[
\sum_{a\in X}\delta_X(a,x) \ge \sum_{a\in X}\delta_Y(a,x)
\]
and
\[
\min_{a\in X}\left( 1 - \mathbf{e}(x)^T \mathbf{e}(a) \right) \ge \min_{a\in Y}\left( 1 - \mathbf{e}(x)^T \mathbf{e}(a) \right).
\]
Moreover, the additional terms from \(Y\setminus X\) in \(\Delta_{F_2}(x\mid Y)\) further decrease the total marginal gain. Thus, we obtain
\[
\Delta_{F_2}(x\mid X) \ge \Delta_{F_2}(x\mid Y).
\]
Hence, \(F_2\) is submodular.

\textbf{(iii) Combination:}  
Since \(F(X; s_t)= \alpha\,F_1(X) + \beta\,F_2(X)\) with \(\alpha,\beta \ge 0\), it follows that for all \(X\subseteq Y\subseteq V\) and for all \(x\in V\setminus Y\):
\[
\begin{aligned}
F(X\cup\{x\}; s_t) - F(X; s_t) &= \alpha\,\Delta F_1(x\mid X) + \beta\,\Delta F_2(x\mid X), \\[1ex]
&\ge \alpha\,\Delta F_1(x\mid Y) + \beta\,\Delta F_2(x\mid Y) = F(Y\cup\{x\}; s_t) - F(Y; s_t).
\end{aligned}
\]
Thus, \(F(X; s_t)\) is submodular.
\end{proof}

\section{Additional Experimental Results and Analysis}\label{sec:appendix_exp}

\subsection{Reasoning Performance Across Difficulty for RQ5}

\begin{table}[h]
\centering
\caption{Evaluation results on the Level 3, Level 4, and Level 5 subsets of MATH-500. Numbers in bold denote the best performance.}
\resizebox{0.4\columnwidth}{!}{
\begin{tabular}{l@{\hspace{0.5em}}cc@{\hspace{1em}}cc@{\hspace{1em}}cc}
\toprule
& \multicolumn{1}{c}{Level 3} & \multicolumn{1}{c}{Level 4} & \multicolumn{1}{c}{Level 5} \\
\midrule
rStar & 72.38 & 50.78 & 15.67 \\ \midrule
\ourmodel & \textbf{76.19} & \textbf{58.59} & \textbf{31.34} \\ 
- util & 68.57 & 52.34 & 17.16 \\
- q-learning & 71.43 & 53.13 & 20.90 \\
\bottomrule
\end{tabular}
}
\label{tab:utility_effectiveness}
\end{table}

We evaluate the effectiveness of both the utility term and the Q-learning objective
using the MATH-500 dataset, which includes problems categorized by difficulty levels. This allows us to assess how each component impacts problem-solving performance across tasks of varying complexity. As shown in Table~\ref{tab:utility_effectiveness}, the removal of the utility term leads to a more significant performance drop on harder problems (Level 5) compared to easier ones (Level 3). This indicates that the utility term is crucial for selecting actions that meaningfully contribute to the solution process, particularly for complex problems. In comparison, removing the Q-learning objective results in a slight performance drop. Without Q-learning, the embedding function \(\mathbf{e}(\cdot)\) still selects actions relevant to the current state, but it fails to capture the long-term utility of those actions. As a result, while the model continues to choose relevant actions, the lack of Q-learning limits its ability to optimize the effectiveness of its actions over time.

We further compare our method with rStar, which demonstrates better performance than other baselines on MATH-500. However, due to rStar’s manually defined action space, it faces challenges in scaling to more complex problems. As shown in the results for Level 5 problems in Table~\ref{tab:utility_effectiveness}, rStar's performance drops significantly to 15.67\%, whereas our method achieves a higher accuracy of 31.34\%. This highlights the scalability advantage of our method.

\subsection{Action Diversity Analysis for RQ6}

\begin{table}[h]
\centering
\caption{Evaluation results showing the impact of the diversity term.}
\label{tab:diversity_effectiveness}
\resizebox{0.3\columnwidth}{!}{
\begin{tabular}{lcc}
\toprule
Model & Diversity & Accuracy \\
\midrule
Ours & 0.73 & 31.34 \\ \midrule
- div & 0.49 & 24.63 \\
\bottomrule
\end{tabular}
}
\end{table}

To evaluate the impact of the diversity term, we follow \citet{wang2024planning} and calculate the diversity score of a candidate action set by measuring how dissimilar the actions are within the set. Specifically, the diversity score is determined by computing the ratio of dissimilar pairs of actions
to the total number of possible action pairs in the set. This ratio is then averaged over all candidate action sets. We test on the Level 5 subset of the MATH-500 dataset, which consists of more complex problems that are sensitive to redundancy in the selected actions. The results are shown in Table~\ref{tab:diversity_effectiveness}. We observe that removing the diversity term results in a significant drop in the diversity score, which subsequently leads to a decrease in accuracy.

\subsection{Additional Analysis on Diversity Term in Submodular Function}

To assess the sensitivity of \textsc{DynaAct} to the choice of the diversity term \( f_{\text{div}} \) in Eq.~\ref{eq:submodular_div}, we conducted additional experiments using two alternative diversity metrics: \textit{Mean Pairwise Distance} and \textit{Mean Cosine Distance}. These alternatives were chosen to evaluate whether simpler diversity formulations could yield comparable or improved performance.

\begin{table}[h]
\centering
\caption{Performance comparison using different diversity terms.}
\label{tab:diversity_analysis}
\begin{tabular}{lc}
\toprule
Method & Accuracy (MATH-500) \\
\midrule
Mean Pairwise Distance & 57.80 \\
Mean Cosine Distance   & 58.20 \\
\ourmodel & \textbf{61.00} \\
\bottomrule
\end{tabular}
\end{table}

As shown in Table~\ref{tab:diversity_analysis}, both alternative diversity metrics lead to a performance drop compared to our original formulation. We attribute this to the fact that these metrics do not preserve the submodular property, which is central to the efficiency and theoretical guarantees of our greedy subset selection algorithm.

\subsection{Resource Consumption Analysis}

While \S\ref{sec_latency_study} presents a relative latency comparison, we include here the raw inference time per example to offer a more complete view of resource consumption. All measurements were taken on an 8$\times$A100 GPU machine using the MATH-500 dataset, with each method evaluated under consistent rollout settings.

\begin{table}[h]
\centering
\caption{Per-example raw inference time and accuracy on MATH-500 across methods.}
\label{tab:resource}
\begin{tabular}{lcc}
\toprule
Method & Raw Time (\(\downarrow\)) & Accuracy (\(\uparrow\)) \\
\midrule
Zero-shot CoT   & 1.68s  & 45.40 \\
SC@maj16        & 26.88s & 52.00 \\
RAP             & 64.51s & 51.60 \\
rStar           & 54.72s & 54.20 \\
\ourmodel & 57.60s & \textbf{61.00} \\
\bottomrule
\end{tabular}
\end{table}

As shown in Table~\ref{tab:resource}, \ourmodel requires comparable runtime to other MCTS-based baselines such as RAP and rStar, while yielding significantly higher accuracy. Although MCTS-based approaches naturally incur more latency than single-pass generation methods like Zero-shot CoT, they also enable more effective reasoning. Since our method is orthogonal to the choice of search algorithm, future work could explore integration with more efficient test-time strategies (e.g., beam search or sample-efficient MCTS variants) to further reduce resource consumption while preserving accuracy.

\subsection{Comparison with Non-Submodular Selection Methods}

To further understand the benefits of our submodular formulation, we compare \ourmodel against a RL-based pruning baseline. In this baseline, action space truncation is performed by selecting the top $5$ candidate actions based solely on Q-value estimates, without considering submodular diversity or joint utility.

\begin{table}[h]
\centering
\caption{Comparison of \ourmodel with an RL-based pruning baseline on the MATH-500 dataset.}
\label{tab:rl_pruning}
\begin{tabular}{lcc}
\toprule
Method & Raw Time (\(\downarrow\)) & Accuracy (\(\uparrow\)) \\
\midrule
RL-based Pruning & 56.89s & 53.20 \\
\ourmodel & 57.60s & \textbf{61.00} \\
\bottomrule
\end{tabular}
\end{table}

As shown in Table~\ref{tab:rl_pruning}, while the RL-based pruning method achieves slightly lower inference latency, it significantly underperforms in accuracy compared to \ourmodel. This gap highlights the limitations of greedy RL-based pruning, which may select redundant or suboptimal actions. In contrast, our submodular approach explicitly optimizes for both utility and diversity, yielding a more compact yet expressive action set.

\subsection{Scalability with Large Proxy Action Spaces}

To assess the scalability of \ourmodel to large-scale corpora, we varied the size of the proxy action space from $40,000$ to $1,000,000$ entries, while keeping the candidate set size $m = 5$ fixed. All experiments were conducted on the MATH-500 dataset.

\begin{table}[h]
\centering
\caption{Scalability analysis of \ourmodel with varying proxy action space sizes on MATH-500.}
\label{tab:scalability}
\begin{tabular}{lcc}
\toprule
Proxy Action Space Size & Raw Time (\(\downarrow\)) & Accuracy (\(\uparrow\)) \\
\midrule
40k   & 57.60s & 61.0 \\
200k  & 60.52s & 61.8 \\
400k  & 63.99s & 62.0 \\
600k  & 67.93s & 62.0 \\
800k  & 71.47s & 61.6 \\
1M    & 75.84s & 61.8 \\
\bottomrule
\end{tabular}
\end{table}

As shown in Table~\ref{tab:scalability}, the inference latency increases moderately with larger proxy spaces—rising by approximately $18$ seconds when scaling from 40k to 1M actions. Importantly, performance remains stable or slightly improves, indicating that \ourmodel effectively handles large candidate pools without significant degradation in efficiency. This scalability is largely attributed to the caching of action embeddings and the linear-time greedy algorithm used in submodular optimization.

\subsection{Empirical Study on the Utility-Diversity Trade-off}

To better understand how the balancing parameters \( \alpha \) and \( \beta \) in Eq.~\ref{eq:submodular} affect the trade-off between utility and diversity in the submodular function, we conducted a series of experiments on the MATH-500 dataset. We varied the relative weighting of the utility term \( f_{\text{util}} \) and the diversity term \( f_{\text{div}} \), while keeping their sum fixed (\( \alpha + \beta = 1 \)).

\begin{table}[h]
\centering
\begin{tabular}{lc}
\toprule
\(\alpha\), \(\beta\) & Accuracy (MATH-500) \\
\midrule
(0.9, 0.1) & 61.00 \\
(0.7, 0.3) & 60.80 \\
(0.5, 0.5) & 55.40 \\
(0.3, 0.7) & 54.60 \\
(0.1, 0.9) & 54.80 \\
\bottomrule
\end{tabular}
\caption{Effect of varying \(\alpha\) and \(\beta\) on model performance.}
\label{tab:alpha_beta}
\end{table}

As shown in Table~\ref{tab:alpha_beta}, performance is highly sensitive to the relative weighting of the utility term. Accuracy drops substantially when utility and diversity are given equal weight or when diversity dominates. However, once the utility coefficient \( \alpha \) exceeds 0.7, the model achieves strong and stable performance. These findings suggest that while diversity contributes to a more robust action space, prioritizing utility is essential for effective reasoning in complex problem-solving tasks.

\subsection{Comparison with Few-Shot and Fine-Tuned Baselines}

To assess whether the performance of \ourmodel could be attributed to exemplar-based prompting or supervised adaptation, we compare it with several baselines involving few-shot prompting and fine-tuning. The few-shot baseline retrieves the top 3 most similar examples from Open-Platypus using cosine similarity over embeddings from Llama-3.2-1B-Instruct, and provides them as context to Llama-3.1-8B-Instruct. The fine-tuned baseline directly trains Llama-3.1-8B-Instruct on the same corpus for 3 epochs with a learning rate of \(1 \times 10^{-5}\). For both approaches, we also evaluate their self-consistency variants (SC@maj16), where 16 reasoning paths are sampled and majority voting is used for answer selection.

\begin{table}[h]
\centering
\caption{Comparison of \ourmodel with few-shot and fine-tuned baselines across six benchmarks.}
\label{tab:fewshot_finetune}
\begin{tabular}{lcccccc}
\toprule
Method & MMLU & MMLU-Pro & GPQA & ARC-C & GSM8K & MATH-500 \\
\midrule
Zero-shot CoT & 68.87 & 43.45 & 31.82 & 81.06 & 76.12 & 45.40 \\
Few-shot baseline & 68.94 & 43.23 & 29.80 & 81.83 & 76.65 & 46.60 \\
Fine-tuned baseline & 68.72 & 43.56 & 34.34 & 82.08 & 78.17 & 48.20 \\
Few-shot SC@maj16 & 69.80 & 44.22 & 35.86 & 84.73 & 84.00 & 50.20 \\
Fine-tuned SC@maj16 & 69.89 & 45.89 & 34.85 & 86.18 & 84.38 & 52.60 \\
\ourmodel & \textbf{70.22} & \textbf{51.40} & \textbf{39.39} & \textbf{88.31} & \textbf{89.16} & \textbf{61.00} \\
\bottomrule
\end{tabular}
\end{table}

As shown in Table~\ref{tab:fewshot_finetune}, both few-shot prompting and fine-tuning lead to modest improvements over the zero-shot baseline. However, \ourmodel consistently outperforms all baselines across tasks, demonstrating that its gains stem from principled, dynamic action space construction rather than static exemplar use or model adaptation alone.

\subsection{Case Study}
\begin{figure}[htb]
\small
    \centering
    \fbox{%
        \begin{minipage}{0.95\textwidth}
            \vspace{0.5em}
            \textbf{Problem:} \\
            A hexagon is inscribed in a circle:
            \begin{center}
            \includegraphics[width=0.3\textwidth]{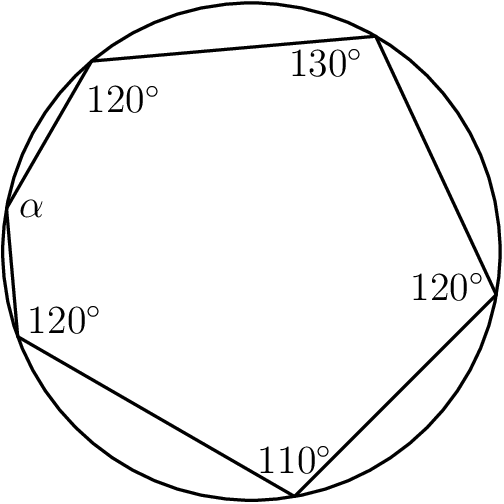} 
            \end{center}
            What is the measure of $\alpha$, in degrees?
            
            \vspace{1em}
            \textbf{Solution:}
            
            \noindent\textbf{Step 1:} I need to identify the underlying structure or pattern in the problem, whether it's a mathematical relationship, a logical argument, or a conceptual framework. \\
            The problem involves an inscribed hexagon in a circle, with specific angles labeled. The goal is to find the measure of angle $\alpha$.
            
            \vspace{0.5em}
            \noindent\textbf{Step 2:} I will decide on a suitable representation or model for the problem, and determine the best approach to explore the solution space, whether it's through calculation, simulation, or other means. \\
            To find the measure of $\alpha$, I can use the fact that the sum of the interior angles of a hexagon is $720^\circ$. Additionally, since the hexagon is inscribed in a circle, the measure of each central angle formed by two adjacent vertices of the hexagon is equal to the measure of the corresponding inscribed angle.
            
            \vspace{0.5em}
            \noindent\textbf{Step 3:} I evaluate my progress, checking for errors, inconsistencies, and areas for improvement, and I adapt my approach as needed to ensure that I'm making progress toward a solution that is accurate, complete, and relevant. \\
            Since the sum of the interior angles of a hexagon is $720^\circ$, and the sum of the measures of the given angles is $105^\circ + 110^\circ = 215^\circ$, the measure of angle $\alpha$ can be found by subtracting the sum of the given angles from the total sum of the interior angles of a hexagon.
            
            \vspace{0.5em}
            \noindent\textbf{Step 4:} I break down complex problems into smaller, manageable parts, focusing on one aspect at a time, and identifying the essential variables, assumptions, and constraints that will help me make progress, which might involve creating a diagram, setting up equations, or developing a step-by-step plan. \\
            Let's denote the measure of angle $\alpha$ as $x$. Since the sum of the interior angles of a hexagon is $720^\circ$, we have the equation $x + 215^\circ = 720^\circ$.
            
            \vspace{0.5em}
            \noindent\textbf{Step 5:} Are there any constraints, assumptions, or rules that I need to take into account when exploring solutions or possibilities? \\
            The problem does not provide any additional constraints or assumptions beyond the given information.
            
            \vspace{0.5em}
            \noindent\textbf{Step 6:} I should look for potential paths forward by asking myself questions like 'What are the implications of this?', 'How does this relate to that?', 'What would happen if I...', and 'What are the boundaries or constraints of this problem?' - this helps me generate possibilities and explore different avenues for solving the problem. \\
            Now, let's solve the equation $x + 215^\circ = 720^\circ$ for $x$.
            
            \vspace{0.5em}
            \noindent\textbf{Step 7:} I break down complex problems into smaller, manageable parts, focusing on one aspect at a time, and identifying the essential variables, assumptions, and constraints that will help me make progress, which might involve creating a diagram, setting up equations, or developing a step-by-step plan. \\
            To solve for $x$, we can subtract $215^\circ$ from both sides of the equation: $x = 720^\circ - 215^\circ$.
            
            \vspace{0.5em}
            \noindent\textbf{Step 8:} Complete the solution and present the final answer within $\boxed{}$. \\
            $x = 720^\circ - 215^\circ = 505^\circ$. However, since the measure of an angle cannot exceed $180^\circ$, we need to subtract $360^\circ$ from $505^\circ$ to get the correct measure of angle $\alpha$. Therefore, $\alpha = 505^\circ - 360^\circ = 145^\circ$.
            \[\boxed{145}\]
        \end{minipage}%
    }
    \caption{Case study: solution to inscribed hexagon angle problem.}
    \label{fig:case_study_3}
\end{figure}

To further demonstrate the effectiveness of \ourmodel, we present a series of case studies in Figures \ref{fig:case_study_1} - \ref{fig:case_study_3}, each consisting of a ``problem'' and its corresponding predicted ``solution''. For each step in the solution, the process begins with the selection of an action~(chosen using $Q(s_t,a)$, as described in \S\ref{sec:reasoning_framework}), followed by the corresponding reasoning step. These examples demonstrate how \ourmodel enhances the effectiveness of the reasoning process.

\begin{figure}[htb]
\small
    \centering
    \fbox{%
        \begin{minipage}{0.95\textwidth}
            \vspace{0.5em}
            \textbf{Problem:} \\
            Compute: $1-2+3-4+5- \dots +99-100$
            
            \vspace{1em}
            \textbf{Solution:}
            
            \noindent\textbf{Step 1:} I clarify the problem by identifying the key elements, constraints, and what I'm trying to find or achieve, and I rephrase it in my own words to ensure I understand it correctly. \\
            We need to compute the sum of an alternating series where each term increases by 1, starting from 1 and ending at 100.
            
            \vspace{0.5em}
            \noindent\textbf{Step 2:} I should look for connections, patterns, or relationships between the components, and consider how they interact or influence each other, whether it's in a mathematical, logical, or contextual sense. \\
            The series alternates between addition and subtraction, which suggests a pattern where each pair of consecutive terms (one positive and one negative) could be combined to simplify the calculation.
            
            \vspace{0.5em}
            \noindent\textbf{Step 3:} I'm searching for patterns, connections, and relationships between the different parts, and thinking about how they might help me solve the problem. \\
            We can pair the terms as follows: $(1-2) + (3-4) + (5-6) + \dots + (99-100)$. Each pair sums to $-1$.
            
            \vspace{0.5em}
            \noindent\textbf{Step 4:} I evaluate my progress, checking if my approach is working and adjusting it as needed to ensure I'm on the right track. \\
            There are 50 pairs in the series, each summing to $-1$.
            
            \vspace{0.5em}
            \noindent\textbf{Step 5:} I break down complex problems into smaller, manageable parts, focusing on one aspect at a time, and identifying the essential variables, assumptions, and constraints that will help me make progress, which might involve creating a diagram, setting up equations, or developing a step-by-step plan. \\
            The sum of the series is the sum of the 50 pairs, which is $50 \times (-1)$.
            
            \vspace{0.5em}
            \noindent\textbf{Step 6:} Complete the solution and present the final answer within $\boxed{}$. \\
            \[\boxed{-50}\]
        \end{minipage}%
    }
    \caption{Case study: solution to alternating series sum.}
    \label{fig:case_study_1}
\end{figure}

\begin{figure}[htb]
\small
    \centering
    \fbox{%
        \begin{minipage}{0.95\textwidth}
            \vspace{0.5em}
            \textbf{Problem:} \\
            The Greek army contained two types of soldiers: the upper class and the lower class soldiers. If there were a total of 5 upper class soldiers, and 10 lower class soldiers in a certain part of Athens, and the battle of Thermopylae demands a force of 4 upper class soldiers and 8 lower class soldiers, how many different battalions can be sent?
            
            \vspace{1em}
            \textbf{Solution:}
            
            \noindent\textbf{Step 1:} I clarify the problem by identifying the key elements, unknowns, and goals, and try to rephrase it in my own words to ensure I understand what's being asked. \\
            We need to find out how many different groups (battalions) can be formed from a mix of 5 upper class soldiers and 10 lower class soldiers, where each group must consist of 4 upper class soldiers and 8 lower class soldiers.
            
            \vspace{0.5em}
            \noindent\textbf{Step 2:} I look for opportunities to use analogies, metaphors, or abstractions to reframe the problem or make it more tractable, and I think about how I can leverage my existing knowledge or experience to inform my solution. \\
            This problem can be approached by considering it as a combinatorics problem, specifically involving combinations since the order of selection does not matter.
            
            \vspace{0.5em}
            \noindent\textbf{Step 3:} I consider the implications and potential consequences of my solution, and I think about how I can test, validate, or refine it further to ensure it's correct, complete, and relevant to the original problem. \\
            To form a battalion, we need 4 upper class soldiers out of 5 and 8 lower class soldiers out of 10. The number of ways to choose 4 upper class soldiers from 5 is given by the combination formula $C(n, k) = \frac{n!}{k!(n-k)!}$, where n is the total number of items, and k is the number of items to choose. For the upper class, this becomes $C(5, 4)$. Similarly, for the lower class, it's $C(10, 8)$.
            
            \vspace{0.5em}
            \noindent\textbf{Step 4:} I consider how the problem might be transformed or re-expressed in a way that makes it more tractable – can I reframe it in simpler terms, or find an analogous problem that's easier to solve, or identify a key insight that unlocks a more straightforward path to the solution? \\
            The formula for combinations is $C(n, k) = \frac{n!}{k!(n-k)!}$. For the upper class, this is $C(5, 4) = \frac{5!}{4!(5-4)!} = 5$. For the lower class, this is $C(10, 8) = \frac{10!}{8!(10-8)!} = 45$.
            
            \vspace{0.5em}
            \noindent\textbf{Step 5:} Are there any constraints, assumptions, or rules that I need to take into account when exploring solutions or possibilities? \\
            Since the selection of upper class soldiers and lower class soldiers is independent of each other, the total number of different battalions that can be sent is the product of the number of ways to choose the upper class soldiers and the number of ways to choose the lower class soldiers.
            
            \vspace{0.5em}
            \noindent\textbf{Step 6:} Once I think I have a solution, I'll consider its implications and potential limitations. I'll ask myself questions like 'What are the potential consequences of this solution?', 'Are there any potential biases or errors in my thinking?', and 'How might this solution impact different stakeholders or scenarios?' \\
            The total number of battalions is the product of the number of ways to choose the upper class and the lower class soldiers, which is $5 \times 45 = 225$.
            
            \vspace{0.5em}
            \noindent\textbf{Step 7:} Look for patterns, symmetries, and analogies that can help me understand the problem more deeply and make connections to other areas of knowledge or experience – are there any clever ways to reframe, reorganize, or reinterpret the problem that could reveal new insights or opportunities? \\
            The solution is straightforward and based on the combination formula.
            
            \vspace{0.5em}
            \noindent\textbf{Step 8:} Complete the solution and present the final answer within $\boxed{}$. \\
            \[\boxed{225}\]
        \end{minipage}%
    }
    \caption{Case study: solution to Greek army battalion formation problem.}
    \label{fig:case_study_2}
\end{figure}

\end{document}